\documentclass[letterpaper]{article} % DO NOT CHANGE THIS
\usepackage{aaai2027}  % DO NOT CHANGE THIS
\usepackage[hyphens]{url}  % DO NOT CHANGE THIS
\usepackage{graphicx} % DO NOT CHANGE THIS
\usepackage{natbib}  % DO NOT CHANGE THIS AND DO NOT ADD ANY OPTIONS TO IT
\usepackage{caption} % DO NOT CHANGE THIS AND DO NOT ADD ANY OPTIONS TO IT
\usepackage[table]{xcolor}
\usepackage{algorithm}
\usepackage{algorithmic}
\usepackage{amsmath}
\usepackage{amssymb}
\usepackage{multirow}
\usepackage{threeparttable}
\usepackage{newfloat}
\usepackage{listings}
\DeclareCaptionStyle{ruled}{labelfont=normalfont,labelsep=colon,strut=off} % DO NOT CHANGE THIS
\floatstyle{ruled}
\newfloat{listing}{tb}{lst}{}
\floatname{listing}{Listing}

\usepackage{booktabs}

\title{\textbf{GARDiff: Graph-Aligned Residual Diffusion for Probabilistic Multivariate Time-Series Forecasting}}
\author{
Rui Han\textsuperscript{\rm 1}, 
Min Yang\textsuperscript{\rm 1}, 
Xu Zhang\textsuperscript{\rm 2}, 
Xinghao Yang\textsuperscript{\rm 3}, 
Wei Liu\textsuperscript{\rm 4},
Yongshun Gong\textsuperscript{\rm 1}
}
\affiliations{
\textsuperscript{\rm 1}Shandong University, Jinan, China\\
\textsuperscript{\rm 2}Macquarie University, Sydney, Australia\\ 
\textsuperscript{\rm 3}China University of Petroleum (East China), Qingdao, China\\
\textsuperscript{\rm 4}University of Technology Sydney, Sydney, Australia\\

\{hanrui021, minyang\}@mail.sdu.edu.cn, xu.zhang12@hdr.mq.edu.au, yangxh@upc.edu.cn, wei.liu@uts.edu.au, ysgong@sdu.edu.cn
}

\begin{document}

\maketitle

\begin{abstract}
    Diffusion models have recently shown strong potential for probabilistic multivariate time-series forecasting by modeling complex conditional distributions. Recent decoupled diffusion frameworks further separate forecasting into deterministic prediction and stochastic residual generation, making it natural to derive dependency graphs from deterministic representations and use them to guide residual diffusion. However, we show that this direct structural transfer is unreliable. Although deterministic-derived graphs encode useful global dependency priors, they exhibit substantial edge-level misalignment with residual dependency structures, introducing inaccurate or redundant conditions during residual generation. This reveals a previously overlooked deterministic-to-residual structural alignment problem in decoupled diffusion forecasting. To address this problem, we propose \textbf{GARDiff}, a \textbf{G}raph-\textbf{A}ligned \textbf{R}esidual \textbf{Diff}usion framework for probabilistic multivariate time-series forecasting. Instead of treating deterministic-derived graphs as fixed diffusion conditions, GARDiff progressively adapts them to residual generation. Specifically, GARDiff estimates residual uncertainty to distinguish high- and low-uncertainty regions, enabling uncertainty-aware structural refinement, and further performs timestep-aware edge sparsification during reverse diffusion to evolve graph conditions from broad dependency aggregation to localized residual refinement. Extensive experiments on six real-world benchmarks demonstrate that GARDiff consistently improves probabilistic forecasting performance and uncertainty calibration over strong baselines.

\end{abstract}

\begin{figure}[h!t]
    \centering
    \includegraphics[width=1\linewidth]{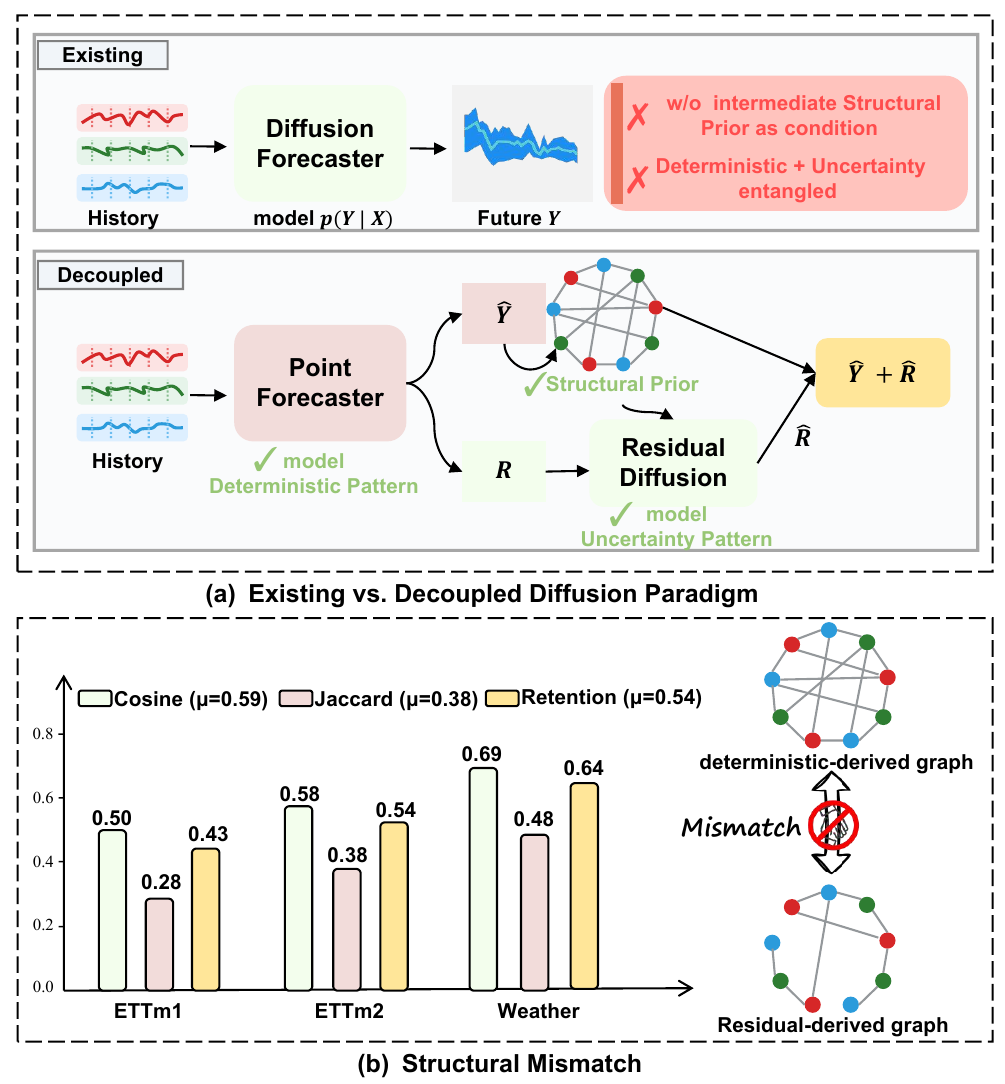}
    \caption{(a) Comparison between existing diffusion forecasting and the decoupled forecasting paradigm. Decoupling deterministic and uncertainty modeling enables the deterministic stage to provide structural priors for uncertainty-aware residual diffusion. (b) Quantitative evidence of structural mismatch between deterministic-derived and residual-derived graphs.}
    \label{fig:motivation}
\end{figure}
\section{Introduction}
Multivariate Time-Series Forecasting (MTSF) plays a critical role in various real-world applications, including energy management~\cite{nowotarski2018recent, deb2017review, xue2023utilizing}, traffic planning~\cite{shu2021short, an2025spatio, li2025adaptive}, weather monitoring~\cite{gong2024spatio, angryk2020multivariate, karevan2020transductive}, urban mobility~\cite{li2024dual, qu2022forecasting, yang2025stda, zhang2023mask, ZHANG2025106900}, and finance~\cite{wiese2020quant, zhu2024lsr}. 
Recently, diffusion models have demonstrated strong potential for probabilistic time-series forecasting due to their capability of modeling complex conditional distributions. However, effective probabilistic forecasting for multivariate time series requires not only modeling temporal uncertainty but also capturing dependencies among variables. Existing graph-conditioned diffusion methods~\cite{wen2023diffstg, liu2023pristi, lin2025specstg} address this challenge by incorporating graph structures as additional priors, yet they typically rely on predefined physical graphs, which are unavailable in many general multivariate time-series scenarios.

Recent decoupled diffusion forecasting frameworks~\cite{li2025diffusion} provide a new perspective by separating forecasting into a deterministic prediction stage and a residual generation stage. As illustrated in Figure~\ref{fig:motivation}(a), this decomposition naturally enables the construction of dependency graphs from deterministic forecasting representations to guide residual diffusion. However, a fundamental question remains: \textbf{can deterministic-derived structures be directly transferred to residual generation?} The deterministic and residual stages optimize different objectives and may capture distinct dependency patterns, making such direct transfer potentially unreliable.

To investigate this issue, we compare deterministic-derived graphs with residual-derived graphs constructed using the same graph generation strategy. As shown in Figure~\ref{fig:motivation}(b), deterministic-derived graphs preserve meaningful global dependency patterns, evidenced by their moderate cosine similarity with residual-derived graphs. However, their substantially lower Jaccard similarity and edge retention ratio reveal significant edge-level inconsistencies between the two stages. This indicates that although deterministic graphs provide valuable structural priors, directly reusing them as fixed conditions may introduce inaccurate or redundant dependencies during residual diffusion. Such structural mismatch is consistently observed across datasets with diverse temporal dynamics, revealing a previously overlooked graph transfer problem in decoupled diffusion forecasting. This motivates a new problem: deterministic-to-residual structural alignment, which aims to adaptively solve the structural mismatch between deterministic forecasting and residual generation while preserving the useful global dependency priors encoded by deterministic-derived graphs.

To address this challenge, structural transfer should be treated as an adaptive alignment process rather than a direct graph reuse strategy. This is because deterministic-to-residual structural mismatch is neither spatially uniform nor temporally static, but varies across residual regions and evolves along the reverse diffusion trajectory. On the one hand, residual uncertainty indicates where richer structural guidance is needed: highly uncertain regions require richer dependency exploration, whereas low-uncertainty regions can benefit from suppressing redundant connections. On the other hand, the reverse diffusion process imposes timestep-dependent structural demands, where early denoising steps favor broader dependency aggregation and later steps require more localized residual refinement. Therefore, effective residual diffusion calls for an alignment mechanism that identifies uncertainty-dependent regions and dynamically adjusts graph edges along denoising timesteps.

Motivated by these observations, we propose \textbf{GARDiff}, a \textbf{G}raph-\textbf{A}ligned \textbf{R}esidual \textbf{Diff}usion framework for probabilistic multivariate time-series forecasting. Instead of using deterministic-derived graphs as fixed diffusion conditions, GARDiff treats them as global structural priors and progressively refines them for residual generation. Specifically, GARDiff first constructs a deterministic dependency graph from intermediate forecasting representations. It then estimates residual uncertainty to distinguish high- and low-uncertainty regions, enabling uncertainty-aware structural refinement. Finally, GARDiff performs timestep-aware edge sparsification during reverse diffusion, allowing the graph condition to evolve from broad dependency aggregation to localized residual refinement. Extensive experiments demonstrate that GARDiff consistently improves probabilistic forecasting performance across multiple benchmarks.

Our main contributions are summarized as follows:
\begin{itemize}
\item We identify and empirically analyze a previously overlooked structural transfer problem in decoupled diffusion forecasting, showing that deterministic-derived graphs provide useful priors but exhibit significant misalignment with residual dependency structures.
\item We propose GARDiff, a graph-aligned residual diffusion framework that couples uncertainty-aware region selection with timestep-aware edge sparsification to align deterministic-derived graph with residual diffusion.
\item Extensive experiments on six real-world benchmarks demonstrate that GARDiff consistently outperforms strong baselines in probabilistic forecasting and uncertainty calibration.
\end{itemize}

\section{Related Work}

\subsection{Diffusion-Based Time-Series Forecasting}

Diffusion probabilistic models~\cite{ho2020denoising} have recently shown strong potential for probabilistic time-series forecasting by modeling complex conditional distributions through iterative denoising. Representative methods adapt diffusion models to sequential forecasting from different perspectives. TimeGrad~\cite{rasul2021autoregressive} performs autoregressive denoising conditioned on historical observations, while CSDI~\cite{tashiro2021csdi} adopts a non-autoregressive Transformer backbone for conditional time-series generation. Subsequent studies further improve temporal representation learning and denoising architectures for forecasting~\cite{alcaraz2022diffusion, li2024transformer}.

Most existing diffusion-based forecasters directly model the future trajectory as a unified stochastic distribution, which entangles predictable temporal patterns with uncertain residual variations. Recent decoupled diffusion frameworks~\cite{li2025diffusion} alleviate this issue by separating deterministic prediction from residual generation. However, they typically use deterministic outputs only as conditioning signals, leaving the dependency structures learned in the deterministic stage underexplored. In contrast, GARDiff extracts deterministic-derived dependency graphs and further studies how such structures should be aligned with residual diffusion.

\subsection{Graph-Conditioned Diffusion Models}

Graph-conditioned diffusion models incorporate graph topology into the denoising process to enhance structured dependency modeling. In spatio-temporal forecasting, DiffSTG~\cite{wen2023diffstg} introduces graph-guided spatial aggregation into diffusion forecasting, PriSTI~\cite{liu2023pristi} extracts coarse spatiotemporal dependencies from observed values as a global context prior and incorporates
geographic relationships into the diffusion-based imputation process, and SpecSTG~\cite{lin2025specstg} extends graph-conditioned diffusion to the spectral domain. These methods demonstrate the effectiveness of graph priors when reliable topology is available.

Nevertheless, existing graph-conditioned diffusion methods are less suitable for general MTSF. They mainly rely on predefined physical graphs, which are unavailable in many non-spatial scenarios, and usually keep the graph condition fixed throughout reverse diffusion. This static design overlooks the evolving structural demands of denoising, where early steps favor broader dependency aggregation while later steps require more localized residual refinement. Different from prior methods, GARDiff derives graphs from deterministic forecasting representations and adaptively refines them through uncertainty-aware region selection and timestep-aware edge sparsification.

\begin{figure*}[h!t]
    \centering
    \includegraphics[width=\textwidth]{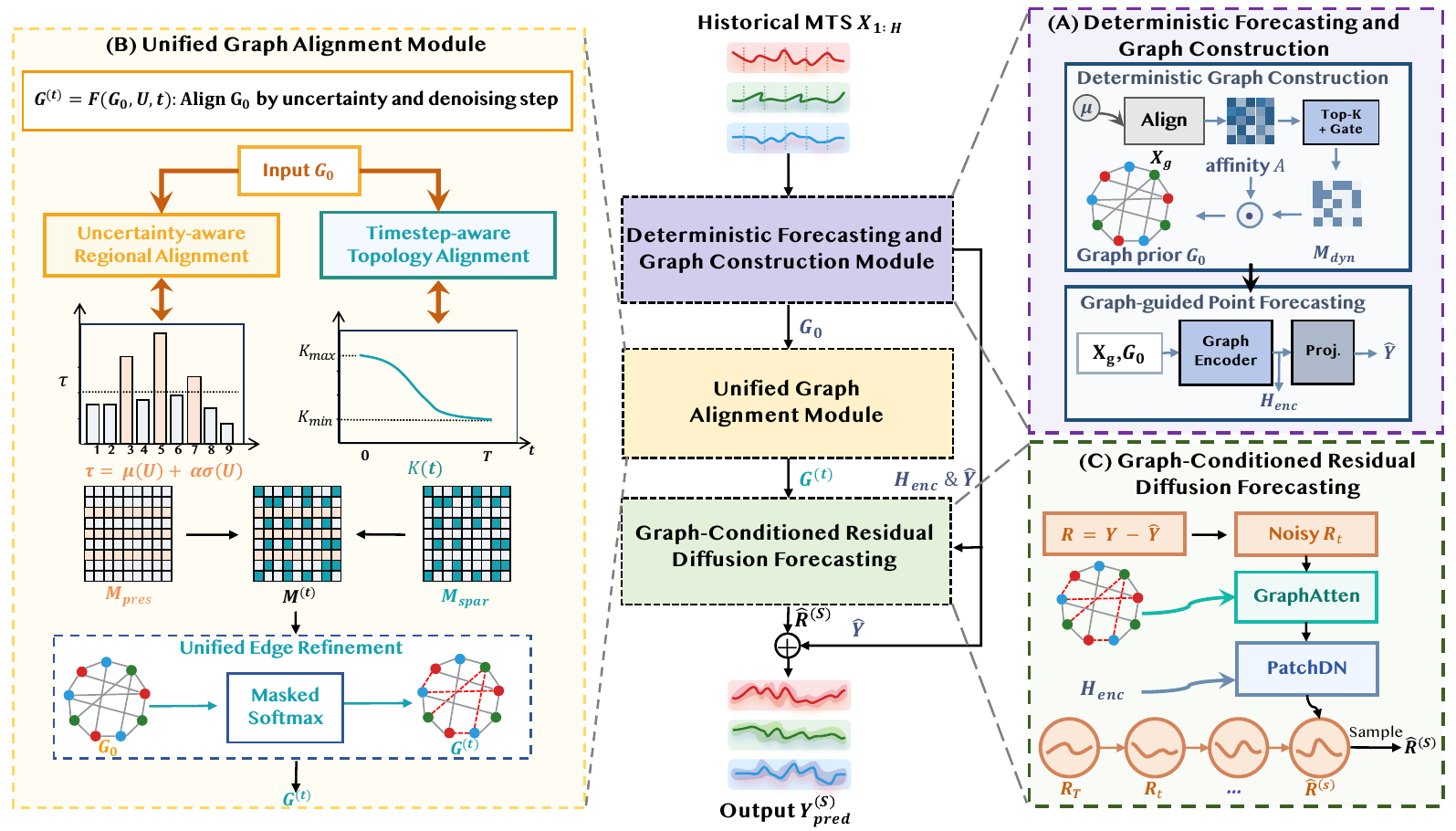}
    \caption{
    Overview of GARDiff. The deterministic forecasting module extracts representations and constructs a structural prior graph. The graph alignment module refines the prior using residual uncertainty and diffusion timestep. The graph-conditioned diffusion module then performs residual generation with dynamic structural guidance.
    }
\end{figure*}
\label{fig:main}

\section{Method}
\subsection{Overview} 
Given historical observations $\mathbf{X}_{1:H}\in\mathbb{R}^{C\times H}$ and prediction target $\mathbf{Y}_{1:L}\in\mathbb{R}^{C\times L}$, GARDiff follows a decoupled forecasting formulation: \begin{equation}
\mathbf{Y}_{1:L}=\hat{\mathbf{Y}}+\mathbf{R}, 
\end{equation}
where $\hat{\mathbf{Y}}=f_{\theta}(\mathbf{X}_{1:H})$ is the deterministic prediction and $\mathbf{R}=\mathbf{Y}_{1:L}-\hat{\mathbf{Y}}$ is the residual. The deterministic stage captures predictable temporal patterns, while the diffusion model is responsible for generating stochastic residuals. 
Different from prior decoupled diffusion methods that mainly use deterministic predictions as conditioning signals, GARDiff further exploits the structural information learned in the deterministic stage. Specifically, we first construct a deterministic dependency graph $\mathbf{G}_0$ from intermediate forecasting representations. Since $\mathbf{G}_0$ is optimized for point prediction rather than residual generation, directly reusing it as a fixed graph condition may introduce residual-irrelevant dependencies. We therefore formulate graph conditioning as an alignment problem: 
\begin{equation}
\mathbf{G}^{(t)}=F(\mathbf{G}_0,\mathbf{U},t), 
\end{equation} where $\mathbf{U}$ denotes patch-level residual uncertainty and $t$ is the reverse diffusion timestep. The aligned graph $\mathbf{G}^{(t)}$ is then used to condition the residual denoising process: 
\begin{equation} 
p_{\phi}(\mathbf{R}\mid \mathbf{H}_{\mathrm{enc}},\mathbf{G}^{(t)}), 
\end{equation} 
where $\mathbf{H}_{\mathrm{enc}}$ denotes deterministic forecasting representations. 

\subsection{Deterministic Forecasting and Graph Construction}

The deterministic forecasting stage serves two purposes: producing the point forecast and extracting a transferable structural prior for residual diffusion. Rather than treating deterministic forecasting solely as a prediction module, we exploit its intermediate representations to model inter-variable dependencies that capture the global dependency backbone of the input series.

\paragraph{\textbf{Patch Embedding.}}
Given an input multivariate time series $\mathbf{X}\in\mathbb{R}^{C\times H}$, we divide each variable into non-overlapping patches of length $p$ and project them into a latent space,
\begin{equation}
\mathbf{X}_p = \mathrm{Embedding}\!\left(\mathrm{Patching}(\mathbf{X})\right)
\in \mathbb{R}^{N\times D},
\end{equation}
where $N=C\times n$, $n=\lceil H/p\rceil$, and $D$ denotes the embedding dimension. Patch-level representations provide stable units for both dependency modeling and deterministic forecasting.

\paragraph{\textbf{Deterministic Graph Construction.}}
To obtain a robust structural prior, we first calibrate patch representations with a global context anchor:
\begin{equation}
\mathbf{X}_{g}=\mathbf{X}_{p}+\mathrm{LN}\!\left(\mathrm{MLP}\left(
[\mathbf{X}_{p}\Vert\boldsymbol{\mu}]\right)\right),
\end{equation}
where $\boldsymbol{\mu}$ is the mean feature over all patches.

Pairwise affinities are then computed by
\begin{equation}
A
=
\mathrm{GELU}
\!\left(
\mathbf{X}_{g}W_1
(\mathbf{X}_{g}W_2)^{\top}
\right),
\end{equation}
followed by Top-$K$ sparsification to retain dominant dependencies. To further suppress noisy structural patterns, we employ a structure-gated mechanism that adaptively reweights candidate dependency structures, yielding the deterministic-derived graph
\begin{equation}
G_0=A\odot M_{\mathrm{dyn}},
\end{equation}
where $M_{\mathrm{dyn}}$ denotes the dynamic structural mask.

\paragraph{\textbf{Graph-guided Point Forecasting.}}
The graph $\mathbf{G}_0$ is used to enhance deterministic forecasting representations: \begin{equation}
\mathbf{H}=\mathrm{GraphEncoder}(\mathbf{X}_{g},G_0),
\qquad\hat{\mathbf{Y}}=\mathrm{Proj}(\mathbf{H}),
\end{equation}
The detailed implementation of the graph encoder and structure gate is provided in the \textbf{Appendix D.1}.
Although $G_0$ captures stable global inter-variable dependencies, it is optimized for deterministic prediction rather than residual generation.  Consequently, directly using $G_0$ as the structural condition for residual diffusion introduces structural bias, motivating the unified graph alignment module described next.

\subsection{Unified Graph Alignment Module}
\label{sec:alignment}

\paragraph{\textbf{Motivation: Refine Rather Than Rebuild.}}
As shown in Figure~\ref{fig:motivation}(b), the deterministic-derived graph $G_0$ preserves a transferable global backbone but differs from the residual-derived graph in fine-grained topology. Thus, $G_0$ is neither fully reliable nor entirely invalid. Discarding it would lose useful structural priors, whereas directly reusing it would introduce residual-incompatible dependencies. We therefore formulate graph conditioning as a \textbf{constrained refinement problem}:
\begin{equation}
    \mathbf{G}^{(t)} = F(G_0,\mathbf{U},t),
\end{equation}
where $\mathbf{U}$ denotes patch-level residual uncertainty and $t$ is the reverse diffusion timestep.

\paragraph{\textbf{Residual Uncertainty Estimation.}}
Graph alignment requires a patch-level uncertainty signal indicating where residual variations are insufficiently explained by the deterministic structure. Under the decoupled formulation $\mathbf{Y}=\hat{\mathbf{Y}}+\mathbf{R}$, the residual $\mathbf{R}=\mathbf{Y}-\hat{\mathbf{Y}}$ represents the stochastic component left for diffusion modeling. Therefore, its magnitude provides a natural proxy for local forecasting difficulty: larger residuals indicate regions where more candidate correction paths should be preserved.

During training, the true residual is directly accessible. During inference, we estimate residual magnitude with a lightweight proxy network:
\begin{equation}
\tilde{\mathbf{R}} = \mathrm{MLP}(\hat{\mathbf{Y}}) \in \mathbb{R}^{C \times L}.
\end{equation}
The proxy is not required to predict the exact residual value, but only to recover the relative residual magnitude used to distinguish high- and low-uncertainty patches, which is substantially simpler than modeling the full residual distribution.

In both training and inference, patch-level uncertainty is computed by reshaping the residual magnitude into patches and aggregating over each patch:
\begin{equation}
\mathbf{U} = \mathrm{Mean}(\mathrm{Patching}({\textit{R}}))
\in \mathbb{R}^{N},
\end{equation}
where $\textit{R}$ denotes the absolute value of $\mathbf{R}$ during training and $\tilde{\mathbf{R}}$ during inference, and $\mathbf{U}_i$ represents the uncertainty of patch $i$.

\paragraph{\textbf{Uncertainty-aware Regional Alignment.}}
To align $G_0$ with residual dependency patterns, we first identify high- and low-uncertainty patches to determine where structural information should be preserved or selectively refined. High-uncertainty patches indicate regions where residual dependencies are less reliably explained by the deterministic backbone. Therefore, aggressively pruning edges in these regions may prematurely remove useful residual-correction paths. In contrast, low-uncertainty patches are better explained by the deterministic model, where redundant connections can be safely suppressed to encourage selective refinement. We do not assume that all edges connected to a high-uncertainty patch are equally important. Instead, high uncertainty indicates a higher risk of removing useful residual-correction paths.

We identify high-uncertainty patches via a dynamic $\alpha\sigma$ 
criterion:
\begin{equation}
    \tau = \mu(\mathbf{U}) + \alpha \cdot \sigma(\mathbf{U}),
\end{equation}
where $\mu(\cdot)$ and $\sigma(\cdot)$ denote the mean and standard 
deviation over all patches, and $\alpha$ is a tunable coefficient. 
Unlike fixed thresholds, this data-adaptive criterion accommodates 
varying uncertainty distributions across datasets and instances. Patches 
satisfying $\mathbf{U}_i > \tau$ are designated as high-uncertainty, for 
which candidate edges are preserved via an uncertainty-preserving mask:
\begin{equation}
M_{\mathrm{pres}}^{ij}
=
\mathbf1
\left[
U_i>\tau
\right],
\end{equation}

\paragraph{\textbf{Timestep-aware Topology Alignment.}}
We dynamically adjust the edge sparsity level along the reverse diffusion trajectory through a Top-$K$ scheduling strategy. In the early reverse diffusion stages (large $t$), 
the denoising network operates under high noise levels and benefits from 
higher-connectivity graph interactions to recover coarse global structure; in later 
stages (small $t$), the signal is nearly clean and localized refinement 
is preferred, making redundant edges harmful rather than helpful. 

We capture this progression via a continuous Top-$K$ scheduling strategy 
with cosine annealing:
\begin{equation}
    K(t) = K_{\min} + \frac{1}{2}(K_{\max} - K_{\min})
    \left(1 - \cos\!\left(\pi \frac{t}{T}\right)\right),
\end{equation}
where $T$ is the total number of diffusion steps, $K_{\max}$ and 
$K_{\min}$ control the connectivity range and $t$ denotes reverse diffusion step. This yields $K(T) = K_{\max}$ at the start of reverse diffusion and $K(0) = K_{\min}$ at convergence, enabling a smooth transition aligned with 
progressive denoising objectives. Furthermore, when $t \geq \eta T$ 
(where $\eta$ is an early-stage threshold), the graph enters a 
maximum-connectivity regime within the candidate graph regime to maximally support coarse structure recovery in 
the noisiest denoising stages. The Top-$K$ sparse mask for low-uncertainty regions is:
\begin{equation}
    M_{\mathrm{spar}} = \mathrm{TopK}(G_0,\ K(t)).
\end{equation}

\paragraph{\textbf{Unified Edge Refinement.}}
The two masks serve complementary roles: $M^{\mathrm{pres}}$ preserves uncertainty-sensitive candidate connectivity irrespective of diffusion progress, while 
$M^{\mathrm{spar}}$ enforces timestep-appropriate sparsity in 
low-uncertainty regions. We unify them via logical disjunction, ensuring 
any edge flagged by either criterion is retained:
\begin{equation}
    M^{(t)} = M_{\mathrm{pres}} \lor M_{\mathrm{spar}}.
\end{equation}
Edges not selected by either mask are suppressed by setting their logits 
to $-\infty$ before normalization. The aligned adjacency matrix is 
obtained via masked softmax:
\begin{equation}
    \mathbf{G}^{(t)} = \mathrm{MaskedSoftmax}(G_0 \odot M^{(t)}).
\end{equation}
Through this unified refinement, $\mathbf{G}^{(t)}$ simultaneously achieves structural alignment with residual dependency patterns and temporal alignment with the evolving objectives of reverse diffusion, providing adaptive 
structural conditioning for residual diffusion.

\subsection{Graph-Conditioned Residual Diffusion}

Given the deterministic prediction $\hat{\mathbf{Y}}$, we model the prediction residual
\begin{equation}
\mathbf{R}=\mathbf{Y}-\hat{\mathbf{Y}}
\end{equation}
using a conditional denoising diffusion process. Rather than learning residual dynamics independently, the diffusion model is conditioned on the aligned graph $\mathbf{G}^{(t)}$, allowing residual generation to exploit dynamically refined structural dependencies throughout reverse diffusion.

\paragraph{\textbf{Forward Diffusion.}}

Following the standard DDPM formulation, Gaussian noise is gradually added to the clean residual:
\begin{equation}
q(\mathbf{R}_t|\mathbf{R})=\mathcal{N}\left(\sqrt{\bar{\alpha}_t}\mathbf{R},
(1-\bar{\alpha}_t)\mathbf{I}\right),
\end{equation}
where
\(\bar{\alpha}_t=\prod_{i=1}^{t}\alpha_i\).
The complete forward and reverse diffusion formulation follows the standard DDPM framework and is omitted for brevity.

\paragraph{\textbf{Graph-Conditioned Denoising.}}

Unlike existing graph-conditioned diffusion models that employ a fixed graph during denoising, our structural condition is dynamically updated at every reverse diffusion step through the graph alignment module:
\begin{equation}
\mathbf{G}^{(t)}=F(\mathbf{G}_0,\mathbf{U},t).
\end{equation}
Consequently, the denoising network receives a timestep-aware structural condition whose connectivity evolves together with the reverse diffusion process, providing dense global interactions in early denoising stages and progressively sparse local refinement as diffusion converges.

\paragraph{\textbf{Graph-Guided Feature Refinement.}}

Before entering the denoising backbone, noisy residual patches
$\mathbf{R}_t\in\mathbb{R}^{N\times P}$
are projected into latent representations with timestep embeddings,
\begin{equation}
\mathbf{E}=\mathrm{Linear}(\mathbf{R}_t)+\mathrm{TimeEmb}(t),
\end{equation}
and refined through a graph-gated multi-head attention module conditioned on
$\mathbf{G}^{(t)}$:
\begin{equation}
\mathbf{E}'=\mathrm{GraphAttn}(\mathbf{E},\mathbf{G}^{(t)}).
\end{equation}

Specifically, the graph is used to modulate self-attention scores through a learnable gating mechanism, enabling structural dependencies to guide feature interaction while preserving the flexibility of Transformer attention. The refined features are added back to the noisy residual through a residual connection before denoising.

\paragraph{\textbf{PatchDN Backbone.}}

The graph-enhanced features are then processed by the PatchDN~\cite{li2025diffusion} denoising backbone to predict either the injected noise or the clean residual, depending on the adopted diffusion parameterization:
\begin{equation}
\hat{\boldsymbol{\epsilon}}_{\phi}
\ \text{or}\
\hat{\mathbf{R}}_{\phi}
=
\mathrm{PatchDN}
(
\mathbf{E}',t,\mathbf{H}_{\mathrm{enc}}
),
\end{equation}
where $\mathbf{H}_{\mathrm{enc}}$ denotes deterministic forecasting representations used as conditional context. PatchDN adopts a standard Transformer architecture with adaptive layer normalization, while the proposed graph-conditioned attention provides dynamic structural guidance throughout reverse diffusion. After $T$ reverse diffusion steps, the denoised residual $\hat{\mathbf{R}}$ 
is added to the deterministic forecast to obtain the final prediction:
\begin{equation}
\mathbf{Y}_{\text{pred}}^{(s)} = \hat{\mathbf{Y}} + \hat{\mathbf{R}}^{(s)}, 
\quad s=1,\dots,S,
\end{equation}
where $s$ denotes the sampling index and $S$ is the total number of samples. 
At inference time, multiple residual samples 
$\{\hat{\mathbf{R}}^{(s)}\}_{s=1}^{S}$ are generated to form 
$\{\mathbf{Y}_{\text{pred}}^{(s)}\}_{s=1}^{S}$, which approximates the predictive 
distribution.

\begin{table*}[t]
  \centering
  {
  \setlength{\tabcolsep}{2.5pt}
  \renewcommand{\arraystretch}{1}
  \begin{tabular}{l|cc|cc|cc|cc|cc|cc}
  \toprule
  \multirow{2}{*}{\textbf{Model}} &
    \multicolumn{2}{c|}{\textbf{ETTm1}} &
    \multicolumn{2}{c|}{\textbf{ETTm2}} &
    \multicolumn{2}{c|}{\textbf{Weather}} &
    \multicolumn{2}{c|}{\textbf{Solar}} &
    \multicolumn{2}{c|}{\textbf{ECL}} &
    \multicolumn{2}{c}{\textbf{Traffic}} \\
  & CRPS & $\mathrm{CRPS}_{\mathrm{sum}}$
  & CRPS & $\mathrm{CRPS}_{\mathrm{sum}}$
  & CRPS & $\mathrm{CRPS}_{\mathrm{sum}}$
  & CRPS & $\mathrm{CRPS}_{\mathrm{sum}}$
  & CRPS & $\mathrm{CRPS}_{\mathrm{sum}}$
  & CRPS & $\mathrm{CRPS}_{\mathrm{sum}}$ \\
  \midrule

  TimeDiff 
  & 0.490 & 2.195
  & 0.320 & 1.597
  & 0.302 & 2.625
  & 0.700 & 2.408
  & 0.735 & 1.966
  & 0.766 & 1.885 \\
  
  \rowcolor{gray!15}
  TMDM 
  & 0.380 & 1.665
  & 0.298 & 1.412
  & 0.244 & 1.915
  & 0.355 & 1.534
  & 0.438 & 1.573
  & 0.453 & 1.587 \\

  NsDiff 
  & 0.394 & 1.878
  & 0.315 & 1.564
  & 0.230 & 1.971
  & 0.273 & 0.951
  & 0.285 & 0.874
  & 0.319 & 0.864 \\
  \rowcolor{gray!15}
  D3U 
  & 0.283 & 0.801
  & 0.247 & 0.114
  & 0.208 & 0.248
  & 0.174 & 0.604
  & 0.202 & 0.665
  & 0.224 & 0.786 \\

  \cmidrule(lr){1-13}

  Ours 
  & \textbf{0.273} & \textbf{0.744}
  & \textbf{0.234} & \textbf{0.093}
  & \textbf{0.199} & \textbf{0.220}
  & \textbf{0.166} & \textbf{0.463}
  & \textbf{0.191} & \textbf{0.459}
  & \textbf{0.212} & \textbf{0.459} \\

  \bottomrule
  \end{tabular}}
  \caption{Probabilistic forecasting performance. All results are reported with input length $H{=}96$ and averaged over three forecasting horizons ($L \in \{96, 192, 336\}$). Best results are highlighted in \textbf{bold}. See Table 6 for full results.}
  \label{tab:prob}
\end{table*}

\begin{table*}[t]
  \centering
  {
  \setlength{\tabcolsep}{5pt}
  \renewcommand{\arraystretch}{1.15}
  \begin{tabular}{l|cc|cc|cc|cc|cc|cc}
  \toprule
  \multirow{2}{*}{\textbf{Model}} &
    \multicolumn{2}{c|}{\textbf{ETTm1}} &
    \multicolumn{2}{c|}{\textbf{ETTm2}} &
    \multicolumn{2}{c|}{\textbf{Weather}} &
    \multicolumn{2}{c|}{\textbf{Solar}} &
    \multicolumn{2}{c|}{\textbf{ECL}} &
    \multicolumn{2}{c}{\textbf{Traffic}} \\
  & MSE & MAE & MSE & MAE & MSE & MAE & MSE & MAE & MSE & MAE & MSE & MAE \\
  \midrule

  \multicolumn{13}{l}{\textbf{Point Forecasting Models}} \\
  \addlinespace[0.3em]

  PatchTST 
  & 0.359 & 0.381 
  & 0.247 & 0.304 
  & 0.227 & 0.258 
  & 0.242 & 0.275 
  & 0.177 & 0.263 
  & 0.455 & 0.287 \\

  \rowcolor{gray!15}
  TimesNet 
  & 0.386 & 0.404 
  & 0.254 & 0.307 
  & 0.225 & 0.263 
  & 0.251 & 0.272 
  & 0.186 & 0.288 
  & 0.613 & 0.325 \\

  iTransformer 
  & 0.381 & 0.397 
  & 0.252 & 0.312 
  & 0.227 & 0.257 
  & 0.231 & 0.257 
  & 0.164 & 0.256 
  & 0.411 & 0.276 \\

  \rowcolor{gray!15}
  TimeBridge 
  & 0.360 & \textbf{0.375} 
  & 0.244 & 0.307 
  & \textbf{0.221} & \textbf{0.248} 
  & 0.220 & \textbf{0.230} 
  & \textbf{0.155} & 0.256 
  & 0.411 & 0.273 \\

  TimeFilter 
  & 0.355 & 0.381 
  & 0.242 & 0.305 
  & 0.226 & 0.258 
  & 0.226 & 0.254 
  & 0.167 & 0.262 
  & \textbf{0.396} & \textbf{0.261} \\

  \midrule

  \multicolumn{13}{l}{\textbf{Probabilistic Forecasting Models}} \\
  \addlinespace[0.3em]

  TimeDiff 
  & 0.573 & 0.503 
  & 0.274 & 0.332 
  & 0.259 & 0.315 
  & 0.901 & 0.714 
  & 0.824 & 0.749 
  & 0.762 & 0.519 \\

  \rowcolor{gray!15}
  TMDM 
  & 0.564 & 0.502 
  & 0.361 & 0.373 
  & 0.287 & 0.303 
  & 0.276 & 0.309 
  & 0.219 & 0.330 
  & 0.625 & 0.374 \\

  NsDiff 
  & 0.561 & 0.489 
  & 0.452 & 0.425 
  & 0.255 & 0.290 
  & 0.307 & 0.329 
  & 0.197 & 0.305 
  & 0.497 & 0.358 \\

  \rowcolor{gray!15}
  D3U 
  & 0.360 & 0.385 
  & 0.247 & 0.315 
  & 0.226 & 0.278 
  & 0.223 & 0.267 
  & 0.179 & 0.269 
  & 0.527 & 0.297 \\

  \cmidrule(lr){1-13}

  Ours 
  & \textbf{0.345} & \textbf{0.375} 
  & \textbf{0.239} & \textbf{0.303} 
  & 0.222 & 0.264 
  & \textbf{0.215} & 0.242 
  & 0.161 & \textbf{0.255} 
  & 0.419 & 0.278 \\

  \bottomrule
  \end{tabular}}
  \caption{Deterministic forecasting performance. All results are reported with input length $H{=}96$ and averaged over three forecasting horizons ($L \in \{96, 192, 336\}$). Best results are highlighted in \textbf{bold}. See Table 7 for full results.}
  \label{tab:deter}
\end{table*}

\paragraph{Optimization Objective.}
GARDiff is trained with a joint objective that combines residual diffusion learning, deterministic forecasting, uncertainty estimation, and structural regularization:
\begin{equation}
\mathcal{L}
=
\mathcal{L}_{\mathrm{diff}}
+
\lambda_1 \mathcal{L}_{\mathrm{point}}
+
\lambda_2 \mathcal{L}_{\mathrm{unc}}
+
\lambda_3 \mathcal{L}_{\mathrm{gate}} .
\end{equation}
Here, $\mathcal{L}_{\mathrm{diff}}$ optimizes residual denoising, 
$\mathcal{L}_{\mathrm{point}}$ supervises the deterministic forecast, 
$\mathcal{L}_{\mathrm{unc}}$ trains the residual uncertainty estimator, 
and $\mathcal{L}_{\mathrm{gate}}$ regularizes the learned graph structure. 
Details are provided in \textbf{Appendix E}.

\section{Experiments}

We evaluate GARDiff on six real-world datasets, focusing on predictive performance and the validity of the proposed structural and uncertainty-aware mechanisms.
\\
\textbf{RQ1:} How does GARDiff perform in probabilistic forecasting compared with state-of-the-art diffusion-based methods? \\
\textbf{RQ2:} Does the proposed framework improve deterministic forecasting accuracy? \\
\textbf{RQ3:} What is the contribution of graph conditioning, topology refinement, and timestep-aware pruning? \\
\textbf{RQ4:} Does the learned uncertainty reliably reflect residual magnitude and support residual-aware graph alignment? \\
\textbf{RQ5:} How sensitive is GARDiff to key hyperparameters?

\subsection{Experimental Setup}

\paragraph{\textbf{Datasets.}}
We evaluate GARDiff on six widely used multivariate time series datasets: ETTm1, ETTm2, Weather, Solar-Energy (Solar), Electricity (ECL), and Traffic. These datasets cover diverse temporal dynamics including strong seasonality and complex inter-variable dependencies.

\paragraph{\textbf{Baselines.}}
We compare GARDiff with representative methods from both deterministic and probabilistic forecasting methods.
(1) Deterministic methods: PatchTST~\cite{nie2022time}, TimesNet~\cite{wu2022timesnet}, iTransformer~\cite{liu2023itransformer}, TimeBridge~\cite{liu2024timebridge}, TimeFilter~\cite{hu2025timefilter}. 
(2) Probabilistic methods: TimeDiff~\cite{shen2023non}, TMDM~\cite{li2024transformer}, NsDiff~\cite{ye2025non}, D3U~\cite{li2025diffusion}. 

\paragraph{\textbf{Evaluation Metrics.}}
For probabilistic forecasting, we use CRPS and CRPS$_{\text{sum}}$. For deterministic forecasting, we report MSE and MAE. Lower values indicate better performance.

\paragraph{\textbf{Implementation Details.}}
We adopt a diffusion process with $T=100$ steps and a linear noise schedule with $\beta_1=10^{-4}$ and $\beta_T=0.02$. For the uncertainty-aware graph alignment module, we set $\alpha=2.0$, and use a continuous Top-$K$ scheduling strategy with $K_{\min}=4$, $K_{\max}=32$, and $\eta=0.75$. During inference, we approximate predictive distributions using 100 diffusion samples. All experiments are implemented in PyTorch~\cite{paszke2019pytorch} and conducted on an NVIDIA GeForce RTX 4090 GPU with 24GB memory. Details are described in \textbf{Appendix F.3}.

\subsection{Overall Results}

\paragraph{\textbf{Probabilistic Forecasting Performance (RQ1).}}
Table~\ref{tab:prob} reports probabilistic forecasting performance in terms of CRPS and CRPS$_{\mathrm{sum}}$.
GARDiff consistently outperforms all competing methods across all six benchmarks.
Compared with the strongest diffusion baseline D3U~\cite{li2025diffusion}, GARDiff achieves up to \textbf{5.5\%} lower CRPS and \textbf{41.6\%} lower CRPS$_{\mathrm{sum}}$.
Such consistent gains suggest that adaptive graph conditioning provides informative structural guidance for residual diffusion and improves distributional calibration across diverse temporal patterns.

\paragraph{\textbf{Deterministic Forecasting Performance (RQ2).}}
Table~\ref{tab:deter} presents deterministic forecasting results.
GARDiff achieves competitive deterministic performance across most benchmarks, particularly ETTm1, ETTm2, and Solar.
Compared with strong baselines including TimeFilter and PatchTST, GARDiff yields consistent improvements in both MSE and MAE.
These results indicate that the learned structural representations are beneficial not only for residual generation but also for deterministic point forecasting.

\begin{figure}[t]
    \centering
    \includegraphics[width=1\linewidth]{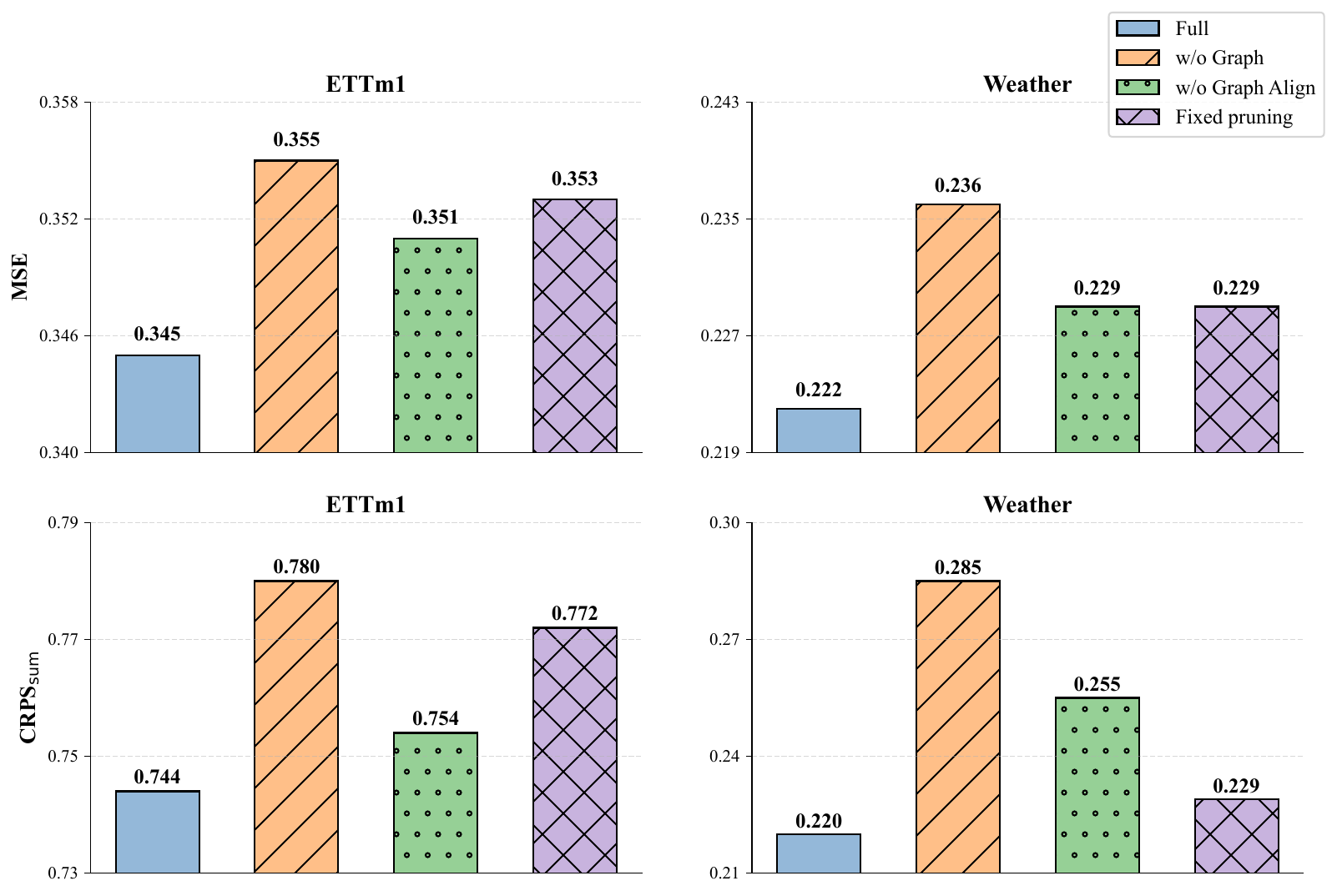}
    \caption{Ablation study on ETTm1 and Weather. The look-back horizon H is 96. See Table 8 for full results.}
    \label{fig:ablation}
\end{figure}

\begin{table}[t]
\centering
\setlength{\tabcolsep}{3pt}
\begin{tabular}{l l c c c c}
\toprule
\textbf{Dataset} & \textbf{Setting} & \textbf{MSE} & \textbf{MAE} & \textbf{CRPS} & $\mathbf{CRPS}_{\mathbf{sum}}$ \\
\midrule
\multirow{2}{*}{ETTm1}
& Pred. Unc. & \textbf{0.345} & \textbf{0.375} & 0.273 & \textbf{0.744} \\
& Oracle Res. & 0.346 & \textbf{0.375} & \textbf{0.272} & 0.751 \\
\midrule
\multirow{2}{*}{Weather}
& Pred. Unc. & \textbf{0.222} & \textbf{0.268} & \textbf{0.202} & \textbf{0.220} \\
& Oracle Res. & \textbf{0.222} & \textbf{0.268} & \textbf{0.202} & \textbf{0.220} \\
\bottomrule
\end{tabular}
\caption{Performance comparison between predicted uncertainty and oracle residuals. Results are reported with input length $H{=}96$ and averaged over three forecasting horizons ($L \in \{96,192,336\}$). See Table 9 for full results.}
\label{tab:uncertainty}
\end{table}

\subsection{Ablation Study (RQ3)}
Three ablated variants are designed to verify the contribution of each core module:
\textbf{-w/o Graph} discards graph conditioning and uses temporal representations only;
\textbf{-w/o Graph Align} directly employs the deterministic graph as a fixed condition without alignment;
\textbf{-Fixed Pruning} applies constant top-8 sparsity across all denoising steps, removing timestep adaptation.

Figure~\ref{fig:ablation} shows consistent performance degradation of all variants on both ETTm1 and Weather datasets.
The drop of \textbf{-w/o Graph} confirms the necessity of graph guidance for capturing inter-variable dependencies.
The degradation of \textbf{-w/o Graph Align} indicates that unrefined deterministic graphs bring suboptimal topology to residual diffusion.
The decline of \textbf{-Fixed Pruning} further demonstrates that static sparsity fails to adapt to evolving denoising objectives.

\subsection{Uncertainty Quality Analysis (RQ4)}
We first evaluate the alignment between estimated uncertainty and oracle residuals via the Exact Matching Rate (EMR), which quantifies the share of patches consistently classified into high/low-uncertainty groups by both criteria. Our estimator achieves EMR of 0.92 on ETTm1 and 0.94 on Weather, verifying its ability to reliably capture local forecasting difficulty.

We further compare downstream forecasting performance using predicted uncertainty versus oracle residuals in Table~\ref{tab:uncertainty}. The two settings yield highly consistent results across all metrics on both datasets. For instance, on 96-step ETTm1, CRPS reaches 0.254 with predicted uncertainty versus 0.250 with oracle residuals. This confirms that the proposed uncertainty proxy acts as a reliable surrogate for oracle residuals without degrading performance.
\begin{figure}[t]
    \centering
    \includegraphics[width=1\linewidth]{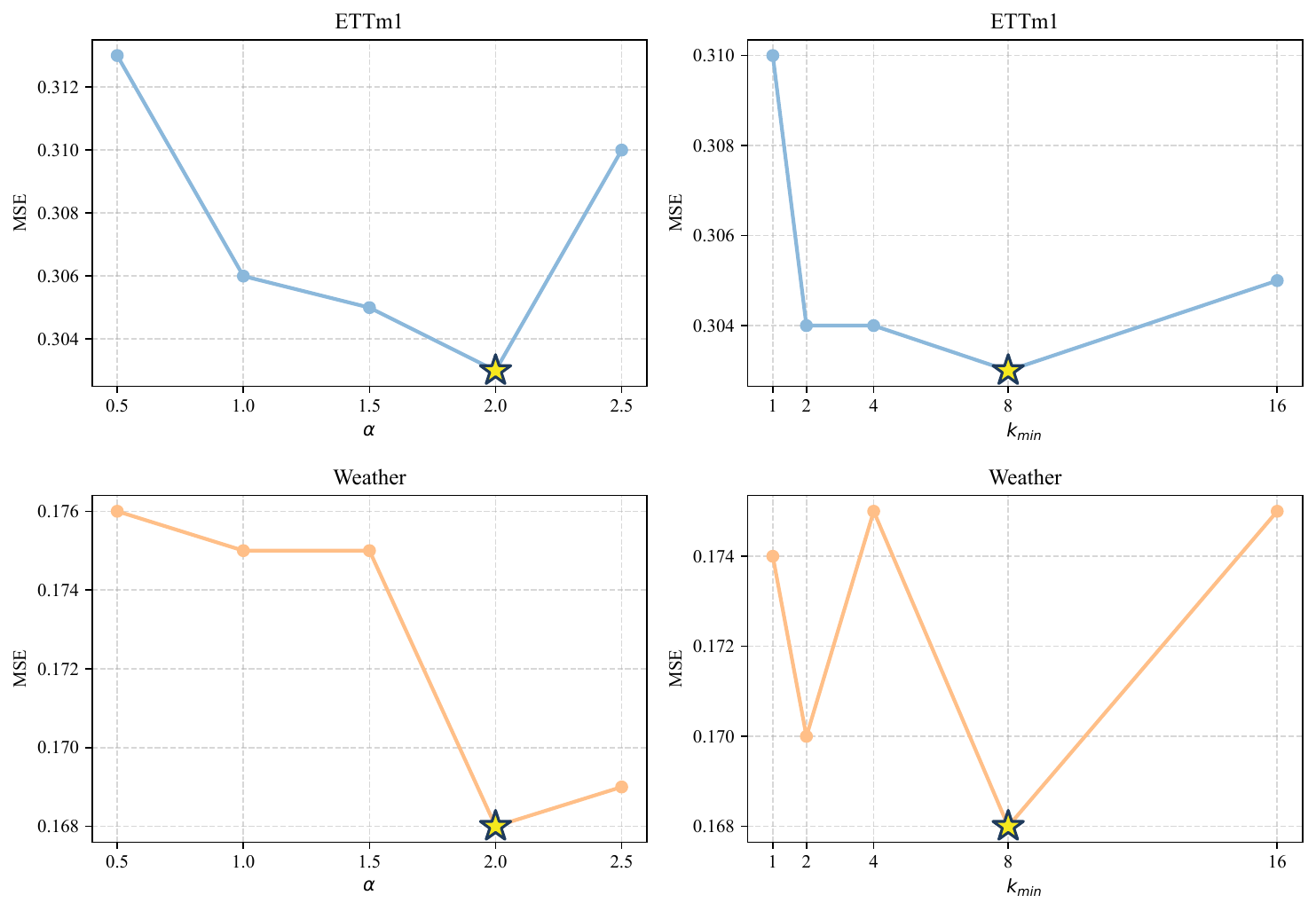}
    \caption{Parameter Sensitivity on ETTm1 and Weather.}
    \label{fig:sensitivity}
\end{figure}
\subsection{Parameter Sensitivity Analysis (RQ5)}
We evaluate the sensitivity of GARDiff to $\alpha$ and $K_{\min}$, which control uncertainty-aware graph construction and sparsity. As shown in Figure~\ref{fig:sensitivity}, the model is stable across a wide range of both parameters, with best performance at moderate values. Too small or too large values of $\alpha$ and $K_{\min}$ both degrade performance due to over-dense or over-sparse structures. Overall, GARDiff is robust to hyperparameter variations.

\section{Conclusion}
In this work, we proposed GARDiff, a Graph-Aligned Residual Diffusion framework for probabilistic multivariate time-series forecasting. Motivated by the structural mismatch between deterministic forecasting and residual generation, GARDiff treats deterministic-derived graphs as global structural priors rather than fixed diffusion conditions. It progressively aligns these priors with residual diffusion through uncertainty-aware regional alignment and timestep-aware topology refinement. The former preserves richer candidate dependencies in high-uncertainty regions, while the latter dynamically adjusts graph sparsity along the reverse denoising trajectory, enabling a transition from broad dependency aggregation to localized residual refinement. Extensive experiments on six real-world benchmarks demonstrate that GARDiff consistently improves probabilistic forecasting performance and uncertainty calibration over strong baselines.

\bibliography{aaai2027}

\end{document}